# Information, Computation, Cognition.
# Agency-based Hierarchies of Levels

Gordana Dodig-Crnkovic

[1] Mälardalen University, Västerås, Sweden
gordana.dodig-crnkovic@mdh.se

**Short Abstract.** Nature can be seen as informational structure with computational dynamics (info-computationalism), where an (info-computational) agent is needed for the potential information of the world to actualize. Starting from the definition of information as the difference in one physical system that makes a difference in another physical system – which combines Bateson and Hewitt's definitions, the argument is advanced for natural computation as a computational model of the dynamics of the physical world where information processing is constantly going on, on a variety of levels of organization. This setting helps elucidating the relationships between computation, information, agency and cognition, within the common conceptual framework, which has special relevance for biology and robotics. (111 words)

**Extended Abstract**

**Nature as Info-computation for a Cognizing Agent**

Naturalizing agency and cognition is possible based on concepts of information and computation as two fundamental and mutually dependent concepts within the framework of info-computationalism. (Dodig-Crnkovic 2009) Information is understood according to informational structural realism (Floridi 2003) (Floridi 2009)(Floridi 2008)(Sayre 1976) as the fabric of the universe (for an agent). Processes of dynamical changes of the universe are interpreted (Zuse 1969, Fredkin 1992, Wolfram 2002, Lloyd 2006, Chaitin 2007) as computation (natural computationalism). Combining the ideas of informationalism and computationalism makes the universe a huge computational network where computation is understood as natural information processing. Natural computation (Rozenberg, Bäck, and Kok 2012) corresponds to the dynamics of processes that exist in the universe, so it necessarily appears in both discrete and continuous form, on both symbolic and sub-symbolic[1] level. A consequence for epistemology for an agent processing information is that *information and reality are seen as one* (Zeilinger 2005) (Vedral 2010), not only for humans, but also for other living organisms as cognizing agents. (Maturana and Varela 1992)(Dodig-Crnkovic 2007) On the info-computational account reality is informational and agent-dependent (observer-dependent) and consists of structural objects integrated into the shared reality of community of practice.

The world exists independently from cognizing agents (realist position of structural realism) in the form of proto-information, the potential form of existence corresponding to Kant's das Ding an sich. That proto-information becomes information ("*a difference that makes a difference*" according to (Bateson 1972)) *for a cognizing agent* in a process of interaction through which aspects of the world get uncovered.

Hewitt proposed the following general relational definition that subsumes Bateson's definition:

"Information expresses the fact that *a system is in a certain configuration that is correlated to the configuration of another system*. Any physical system may contain information about another physical system." (Hewitt 2007) (italics added)

This has consequences for epistemology and relates to the ideas of participatory universe (Wheeler 1990), endophysics (Rössler 1998) and observer-dependent knowledge production of second-order cybernetics. Combining Bateson and Hewitt insights, on the basic level, *information is the difference in one physical system that makes a difference in another physical system, thus constituting correlation between their configurations*.

Among correlated systems, of special interest in our discussion of naturalized cognition are *agents - systems able to act on their own behalf*.

The world as it appears to cognizing agents[2] depends on the types of interactions through which they acquire information[3]. Potential information in the world is obviously much richer than what we observe, containing invisible worlds of molecules, atoms and sub-atomic phenomena, distant objects and similar. Our knowledge about this potential information which reveals with help of scientific instruments continuously increases with the development of new devices and the new ways of interaction with the world, through both theoretical and material constructs (Dodig-Crnkovic and Mueller 2009).

---

[1] Sub-symbolic computations go on in neural networks, as signal processing.

[2] Agents can be any living organisms, plants, animals and humans but even artificial agents.

[3] For example, results of observations of the same physical object (celestial body) in different wavelengths (radio, microwave, infrared, visible, ultraviolet and X-ray) give profoundly different pictures.

This article provides arguments that the new kind of understanding of lawfulness in the organization of nature and especially living systems rests on generative computational laws of *self-organization based on the concept of agents*. In order to understand the world, organization of the parts in the wholes and interactions between them are crucial. That is where generative processes such as self-organization (Kauffman 1993), autopoiesis (Maturana & Varela 1980) and similar come in and they can be tackled by agent-based system-level models such as actor model of computation (Hewitt 2012). The current models of parallel computation (including Boolean networks, Petri nets, Interacting state machines, Process calculi etc.) need to adjust to modelling of biological systems (Fisher and Henzinger 2007). Computational approaches enable a two-way learning process between biology and computing will help formalizing biology as a complex network of dynamical informational structures.

The basic claim of this article is that *nature computes by information processing going on in networks of agents, hierarchically organized in layers*. Informational structures self-organize through processes of natural/physical/embodied computation. (Dodig-Crnkovic and Giovagnoli, 2013)

As an example, newly started Origins *of Life project* lead by Kauffman (University of Vermont) and Markus (CERN) can be mentioned. In their Brainstorming Workshop in 2011 it was announced that the origins of life might be reconstructed within ten years and first living cells produced from purely chemical components. *That indicates the level of our present day knowledge about the concrete mechanisms that drive morphogenetic computing in the physical world of cells*. But there are numerous pieces of insights already now and as our tools are information and computation, info-computational approach offers a suitable unified framework to address form generation and related questions such as emerging in origins of life and nanotechnology.

However, if we want tools to manipulate physical systems, such as molecules in the case of studies of origins of life, our tools must be more than theoretical models – they will be *computations "in materio"* as Stepny (Stepney 2008) called them, arguing aptly: "We are still learning how to use all those tools, both mathematical models of dynamical systems and executable computational models and currently developing 'computation in materio'." That is the reason why physical/natural computing is so important.

In the rest of the article, the basic claims made in the above will be explicated, in the following chapters:

**The Hierarchy of Levels of Physical Computation**

Based on (Burgin and Dodig-Crnkovic 2011) types of computation are systematized to emphasize complexity as fundamental for life and thus for computational modeling of cognizing agents.

**Info-computationalist Constructivist Epistemology for Cognizing Agents**

Elaborates epistemological questions of info-computational agent-based approach in the sense of (Piaget 1955) "Intelligence organizes the world by organizing itself." based on Maturana and Varela's equating life and cognition (Maturana and Varela 1980). Reality as simulation (Smolensky and Legendre 2006), (Sloman 1996)

**System vs. Environment: Open vs. Closed. Self-organization**

Cognition as morphological computation in an agent interacting with the environment – necessity of openness. (Burgin and Dodig-Crnkovic 2013)

**Conclusions**

Summarizes, explaining why understanding of agents (structural and dynamical) is necessary for our understanding of the world as reality which depends on the type of cognizing system that re-constructs it within its own cognitive system.